%% file: sec0_main.tex
\documentclass[sigconf]{acmart}

\AtBeginDocument{%
  }

\setcopyright{acmlicensed}
\copyrightyear{2018}
\acmYear{2018}
\acmDOI{XXXXXXX.XXXXXXX}
\acmConference[Conference acronym 'XX]{Make sure to enter the correct
  conference title from your rights confirmation emai}{June 03--05,
  2018}{Woodstock, NY}
\acmISBN{978-1-4503-XXXX-X/18/06}

\setcopyright{none}

\acmConference{}{}{}
\acmISBN{}
\acmDOI{}
\renewcommand\footnotetextcopyrightpermission[1]{}
\usepackage{enumitem}
\usepackage{booktabs}
\usepackage{multirow}
\usepackage{xcolor}
\usepackage{tabularx}
\usepackage{geometry}
\usepackage{booktabs}
\usepackage{caption}
\usepackage{adjustbox}
\usepackage{float}
\usepackage{multirow}
\usepackage[skip=2pt]{caption} 
\usepackage{graphicx}
\usepackage{colortbl}
\usepackage{ulem}
\usepackage{subcaption}
\usepackage{array}
\usepackage{algorithm}     
\usepackage{algorithmic}
\usepackage{makecell}

\begin{document}

\title{AR-Med: Automated Relevance Enhancement in Medical Search via LLM-Driven Information Augmentation}

\author{%
{\bf Chuyue Wang$^1$, Jie Feng$^2$, Yuxi Wu$^3$, Hang Zhang$^1$, Zhiguo Fan$^1$, Bing Cheng$^1$, Wei Lin$^1$} \\
$^1$Meituan Inc., Beijing, China \\
$^2$Tsinghua University, Beijing, China \\
$^3$Independent Researcher, Beijing, China \\
\texttt{fengj12ee@hotmail.com}
}

\renewcommand{\shortauthors}{Chuyue Wang, et al.}

\begin{abstract}
  Accurate and reliable search on online healthcare platforms is critical for user safety and service efficacy. Traditional methods, however, often fail to comprehend complex and nuanced user queries, limiting their effectiveness. Large language models (LLMs) present a promising solution, offering powerful semantic understanding to bridge this gap. Despite their potential, deploying LLMs in this high-stakes domain is fraught with challenges, including factual hallucinations, specialized knowledge gaps, and high operational costs. To overcome these barriers, we introduce \textbf{AR-Med}, a novel framework for \textbf{A}utomated \textbf{R}elevance assessment for \textbf{Med}ical search that has been successfully deployed at scale on the Online Medical Delivery Platforms. AR-Med grounds LLM reasoning in verified medical knowledge through a retrieval-augmented approach, ensuring high accuracy and reliability. To enable efficient online service, we design a practical knowledge distillation scheme that compresses large teacher models into compact yet powerful student models. We also introduce LocalQSMed, a multi-expert annotated benchmark developed to guide model iteration and ensure strong alignment between offline and online performance. Extensive experiments show AR-Med achieves an offline accuracy of over 93\%, a 24\% absolute improvement over the original online system, and delivers significant gains in online relevance and user satisfaction. Our work presents a practical and scalable blueprint for developing trustworthy, LLM-powered systems in real-world healthcare applications.
\end{abstract}

\maketitle

\input{sec1_intro}
\input{sec2_methods}

\input{sec3_experiments}

\input{sec4_relatedwork}

\section{Security and Ethical Considerations}
Deploying LLM-powered systems in high-stakes domains like medical search demands a rigorous approach to security, ethics, and trustworthiness. Our AR-Med framework incorporates several mechanisms designed to address these challenges, ensuring the system operates safely and responsibly.

Trustworthy Data Traceability. A primary ethical concern with LLMs is their potential for "hallucination," which is unacceptable in a medical context. Our Retrieval-Augmented framework directly mitigates this risk by grounding the model's reasoning in a corpus of verified, professional medical knowledge. Every piece of information used for relevance assessment is retrieved from a trusted source, making the decision process traceable and auditable. This ensures that the system's outputs are not arbitrary but are anchored in reliable data, providing a clear path for verification and accountability.

Continuous Monitoring of Harmful Cases. To prevent patient harm from incorrect or irrelevant search results, we have established a continuous monitoring system. As described in the construction of our LocalQSMed benchmark, we actively identify, sample, and analyze "bad cases"—instances of model failure, such as relevance misjudgment or vulnerability to misleading merchant information (e.g., SPU cheating). These cases are systematically logged and reviewed by medical experts. This feedback loop allows us to rapidly detect and rectify potential safety issues, forming a critical component of our risk management strategy.

Human-in-the-Loop via Expert Rule Governance. Recognizing that automated systems cannot foresee all edge cases, AR-Med integrates a "human-in-the-loop" governance model through its expert rule system (Section 2.2.2). This allows medical professionals to dynamically update and refine the system's decision logic. For example, experts can inject rules to handle newly approved drugs, address emerging public health concerns, or explicitly forbid dangerous product recommendations for specific queries. This expert-driven oversight ensures that the system's behavior remains aligned with the latest medical standards and ethical guidelines, providing an essential layer of safety and control.

Through these integrated strategies, AR-Med provides a blueprint for developing not only an effective but also a trustworthy and ethically-sound AI system for real-world healthcare applications.

\section{Conclusion}
We proposed AR-Med for relevance judgment in pharmaceutical search and recommendation, and constructed a benchmark for pharmaceutical relevance evaluation. Experimental results show that the overall precision on the benchmark improved from 69\% to 93\%, effectively addressing some of the bad cases present in the previous system and bringing measurable benefits after deployment. In the future, our work will focus on further improving the precision of "less relevant" and "irrelevant" categories across different product types, building benchmarks for multiple iterations, and gradually reducing bad cases in Online Medical Delivery’s online system.

\bibliographystyle{ACM-Reference-Format}
\bibliography{sample-base}

\input{sec5_appendix}

\end{document}

%% file: sec1_intro.tex
\section{Introduction}

In the realm of healthcare, accurate pharmaceutical search and recommendation are paramount, as they directly impact public health and safety. With the proliferation of online platforms, medications and medical products are readily available for purchase anytime and anywhere, heightening the need for precise relevance matching to prevent misuse, adverse effects, or ineffective treatments.

Traditional search and recommendation methods, primarily based on small-scale deep learning models, face significant limitations in pharmaceutical domains. These methods include collaborative filtering (e.g., Slope One, user/item similarity \cite{10.1155/2009/421425, 10.1145/3372454.3372470}), content-based filtering (e.g., keyword matching, semantic analysis \cite{10.5555/1394399}), and hybrid approaches like knowledge graph-based systems \cite{guo2020surveyknowledgegraphbasedrecommender}. They struggle with: (1) limited comprehension, failing to handle ambiguous queries like "wind-cold common cold" or "001" \cite{10.1007/978-3-030-99736-6_9}; (2) reliance on costly expert-curated knowledge, increasing maintenance costs \cite{10.5555/1892099}; and (3) slow adaptation to new drugs, regulations, and user demands, leading to outdated relevance \cite{10.5555/1941884}. These challenges highlight the need for more robust solutions.

Large language models (LLMs), on the other hand, hold immense potential to revolutionize this field through their advanced natural language understanding, commonsense reasoning, and zero/few-shot learning abilities~\cite{touvron2023llama, ChatGPT, wei2022emergent}. These capabilities enable LLMs to delve deeper into user query intents and the intrinsic logic of medical texts, thereby substantially enhancing the quality of relevance assessments. For instance, in the domain of natural language understanding, models like GPT-4 have demonstrated expert-level performance on medical question-answering tasks, such as diagnosing complex cases or interpreting clinical queries, by leveraging vast pre-trained knowledge to achieve accuracies comparable to human specialists on benchmarks like USMLE \cite{nori2023capabilitiesgpt4medicalchallenge}. In commonsense reasoning, techniques such as chain-of-thought prompting allow LLMs to break down multi-step problems, improving arithmetic, symbolic reasoning, and logical inference in diverse scenarios, as shown in studies on PaLM models \cite{10.5555/3600270.3602070}. Zero/few-shot learning further empowers LLMs to generalize to unseen tasks with minimal examples; for example, models like GPT-3 excel in translation, code generation, and multitask adaptation without task-specific training \cite{Radford2019LanguageMA}, while specialized variants like CodeGen handle program synthesis in coding domains \cite{nijkamp2023codegenopenlargelanguage}. Across various fields~\cite{Zhang_2024, wang2023huatuotuningllamamodel, Scao2022BLOOMA1, feng2024citygpt, feng2024agentmove, lan2025open, lan2025benchmarking}, LLMs have shown versatility: in e-commerce search, they enhance relevance judgments for product queries \cite{10.1007/978-3-031-56066-8_1}; in biomedical applications, models like HuaTuo and BiomedGPT fine-tuned on Chinese medical knowledge enable accurate diagnosis and knowledge retrieval \cite{wang2023huatuotuningllamamodel, Zhang_2024}; and in tool-augmented settings, Toolformer self-teaches API usage for tasks like calculation or search integration \cite{10.5555/3666122.3669119}. These examples illustrate how LLMs' emergent abilities, such as those in BLOOM's multilingual processing \cite{Scao2022BLOOMA1} or Llama 3's broad cognitive tasks \cite{grattafiori2024llama3herdmodels}, drive innovations in knowledge-intensive applications.

\begin{figure*}[t]
    \centering
    \includegraphics[width=1.0\textwidth]{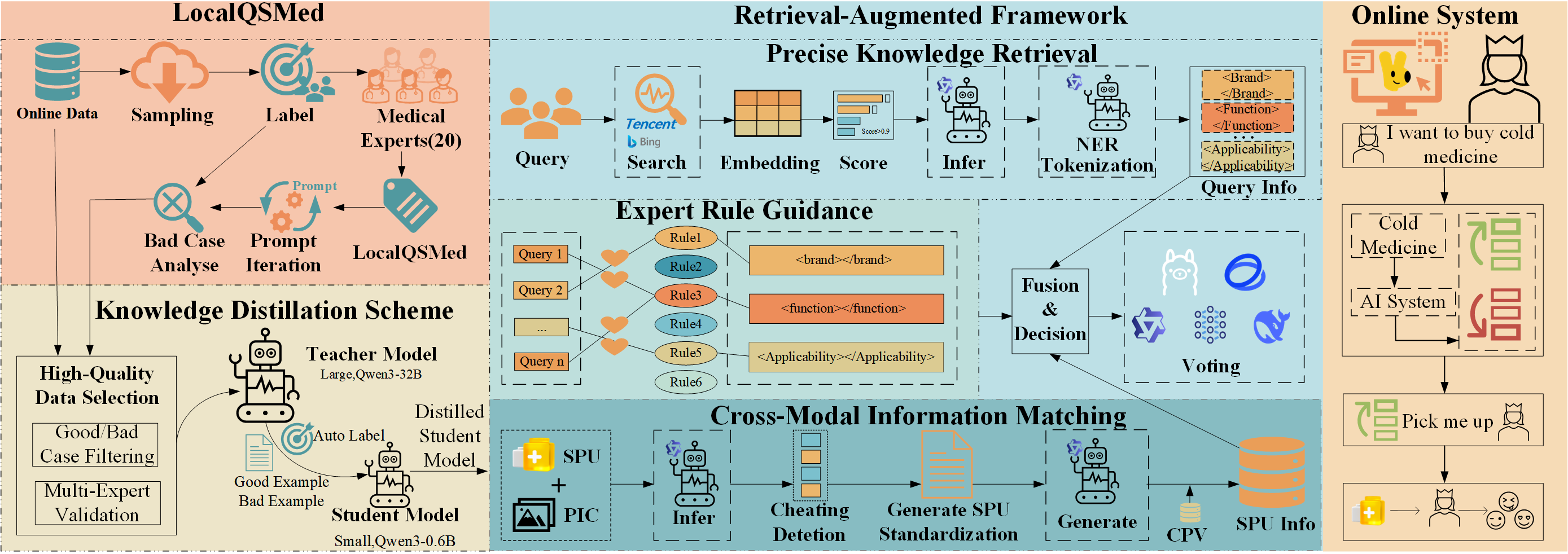}
    \caption{An illustration of the proposed framework, encompassing three key components: Knowledge Retrieval, Expert Rule Guidance, and Cross-modal Information Matching.}
    \label{fig:main}
\end{figure*}

However, LLMs are not without drawbacks: they often lack domain-specific professional knowledge, are prone to severe hallucinations that can lead to erroneous medical advice, and impose high application costs in terms of latency and computational resources, particularly in high-throughput industrial scenarios like daily billions of search requests on platforms such as the Online Medical Delivery Platforms, a leading local services provider. For example, despite strong general capabilities, LLMs like GPT-4 may produce inaccurate outputs in specialized medical contexts due to insufficient domain grounding, as evidenced in evaluations showing persistent errors in clinical reasoning \cite{liévin2023largelanguagemodelsreason}. Hallucinations—fabricating plausible but false information—remain a critical issue, with surveys highlighting risks in high-stakes domains like healthcare \cite{zhao2025surveylargelanguagemodels, Zhou2023DontMY}. 
Moreover, the enormous scale of models such as PaLM 2 (up to 540B parameters) \cite{10.5555/3648699.3648939} or GPT-NeoX-20B \cite{black-etal-2022-gpt} demands massive compute resources for training and inference, leading to inefficiencies in deployment, as noted in compute-optimal analyses \cite{10.5555/3600270.3602446} and efforts to mitigate via optimizations like DeepSpeed \cite{10.1145/3394486.3406703}. These limitations underscore the need for hybrid approaches, such as retrieval-augmented generation (RAG) \cite{10.5555/3495724.3496517}, to complement LLMs in practical settings.

To harness the semantic understanding and reasoning strengths of LLMs while mitigating their weaknesses—namely, insufficient specialized knowledge and hallucination risks—we propose a comprehensive LLM-based framework tailored for medical search and recommendation relevance. Building on the Online Medical Delivery Platforms's diverse and complex real-world scenarios, our approach addresses these challenges through three key innovations. First, we construct a retrieval-augmented framework that deeply integrates LLMs' reasoning capabilities with external, trustworthy medical knowledge bases. This framework employs precise knowledge retrieval, expert-guided rules, and cross-modal information matching to furnish LLMs with verified contextual information for decision-making, thereby anchoring outputs to ensure professionalism, accuracy, and reduced risks in medical relevance judgments. Second, we design a knowledge distillation scheme from large teacher models to compact student models, incorporating offline accumulation of high-confidence online data and targeted distillation learning strategies, to achieve efficient deployment without sacrificing performance. Third, we introduce LocalQSMed benchmark, a comprehensive offline benchmark annotated by multiple experts, encompassing user search queries and product names. This benchmark guides iterative model and system updates offline, providing reliable references for online service enhancements and enabling correct, data-driven iterations. 
In summary, our contributions are as follows:
\begin{itemize}[leftmargin=1.5em,itemsep=0pt,parsep=0.2em,topsep=0.0em,partopsep=0.0em]
\item We propose AR-Med, a novel framework for LLM-based automatic relevance assessment in medical search and recommendation, which integrates large language models with professional medical knowledge to achieve reliable, generalizable, and efficient relevance enhancement for medical search.
\item We design a retrieval-augmented framework that fuses LLMs' reasoning with external medical knowledge bases via precise retrieval, expert rules, and cross-modal matching, anchoring outputs for professional accuracy.
\item We propose a knowledge distillation framework for efficient teacher-to-student model transfer, enabling high-performance online inference. Besides, a multi-expert annotated benchmark is introduced to evaluate offline progress, ensure consistency with online results, and guide effective model iteration.
\item Extensive validations on the Online Medical Delivery Platforms's show over 93\% offline accuracy (24\% above the 69\% baseline), with significant gains in online relevance, user satisfaction, and efficiency.
\end{itemize}

%% file: sec2_methods.tex
\section{Methods}
Our AR-Med approach comprises three primary components, as illustrated in Figure~\ref{fig:main}. First, we introduce a retrieval-augmented generation framework that integrates LLMs' reasoning capabilities with external medical knowledge, expert rules, and multimodal verification to enhance accuracy and mitigate hallucinations. Second, we implement a knowledge distillation scheme leveraging accumulated real-world data to transfer expertise from large teacher models to efficient student models, ensuring cost-effective deployment. The subsequent sections elaborate on each component in detail. Finally, we establish the LocalQSMed benchmark to facilitate systematic offline evaluation and iterative optimization of pharmaceutical relevance assessment. We formalize the complete hierarchical decision logic—integrating precise knowledge retrieval, expert rule guidance, and fine-grained relevance discrimination—in Algorithm~\ref{alg:relevance_judge}. 
Our work presents a practical and scalable blueprint for developing trustworthy, LLM-powered systems in real-world healthcare applications. The detailed specifications of the models employed in our framework, including their specific functions and input/output formats, are summarized in Table~\ref{tab:model_roles}.

\begin{table*}[t]
\centering
\caption{Default models and their roles in the AR-Med framework}
\label{tab:model_roles}
\setlength{\tabcolsep}{2pt}
\footnotesize                         %
\resizebox{1\textwidth}{!}{
\begin{tabular}{@{}l l l p{4.8cm} p{3.8cm}@{}}
\toprule
\textbf{Component} & \textbf{Model}            & \textbf{Function}                     & \textbf{Input}                                      & \textbf{Output} \\
\midrule
\multirow{3}{*}{Query}
   & Qwen3-0.6B-emb            & Similarity Calculation                & Query + web info (e.g., ``001'' $\to$ car model) & Similarity score (e.g., 0.603) \\
   & Qwen3-0.6B (SFT)          & Consistency Judgment                  & Query + web info                                   & Consistency result (e.g., inconsistent) \\
   & Qwen3-32B                 & NER                                   & Filtered query + extended info + rules             & Entities (brand, efficacy, category, etc.) \\
\midrule
\multirow{3}{*}{SPU}
   & Qwen2.5-7B-vl             & Detect Cheating in Product Names      & SPU name + product images                          & Cheating flag (e.g., traffic hijacking) \\
   & Qwen3-32B                 & Standardize Product Name              & Original SPU + cheating result                     & Clean standardized SPU name \\
   & Qwen3-32B + Qwen2.5-7B-vl  & Extend Product Information            & Std. SPU + images + CPV info                       & Rich attributes (efficacy, usage, etc.) \\
\bottomrule
\end{tabular}}
\end{table*}

\subsection{Preliminaries}
Here, we present the fundamental definitions of the core entities in our problem space.

\begin{itemize}[leftmargin=1.5em,itemsep=0pt,parsep=0.2em,topsep=0.0em,partopsep=0.0em]
\item \textbf{Query}: user's raw search input. These are derived from authentic user behavior and can be highly varied, ranging from specific product names (e.g., "Ganmaoling Granules"), to product codes ("001"), or general needs ("anti-dandruff shampoo").

\item \textbf{Standard Product Unit (SPU)}: a standardized product entry in our catalog, sourced from merchant-uploaded information. Our product catalog is extensive, encompassing categories such as Chinese and Western medicines, nutritional supplements, and health-focused personal care products.

\end{itemize}

\subsection{Retrieval-Augmented Framework}
To leverage LLMs' semantic understanding while addressing their limitations in domain knowledge and hallucinations, we design a RAG framework that fuses LLMs with verified external contexts. This framework operates end-to-end, optimizing inputs from both query and SPU sides via precise knowledge retrieval, expert-guided rules, and cross-modal matching, thereby anchoring LLM outputs for professional, accurate relevance judgments.

\begin{table}[htbp]
\centering
\caption{Bad cases caused by overthinking}
\begin{tabularx}{0.48\textwidth}{|l|X|}
\hline
\textbf{prompt} & 
You are an expert in relevance judgment...

\textcolor{blue}{User query}: PiKangWang.

User query interpretation result: ...

\textcolor{blue}{Product SPU}: [PiKangWang] Clotrimazole Cream 10g/bottle Baibang Qiling Antibacterial Ointment.

Product SPU extended information: ...

Basic principles: ...

The output result format is: <result>xxx</result> \\
\hline
\textbf{output} & 
<think>Okay, I now need to...

According to the rules, if the product name contains the brand from the query and the efficacy is similar, it is considered highly relevant. \textcolor{red}{However, in this case, the product's efficacy is antibacterial, while the user's query is for antifungal, which may belong to different categories.}

Therefore, it may be judged as low relevance. The final conclusion may be low relevance.</think> \\
\hline
\end{tabularx}
\vspace{-10pt}
\label{tab:table1}
\end{table}

\subsubsection{Precise Knowledge Retrieval}
In practical settings, user queries are typically concise and multifaceted, complicating relevance assessment with generic LLMs alone. Traditional methods rely on query segmentation and weighted intent matching, but LLMs offer superior natural language comprehension. However, in medicine, abbreviations and jargon (e.g., "wind-cold common cold") can lead to incomplete expansions if relying solely on LLMs, potentially missing key intents like associated medication names.

To mitigate this, we implement a two-stage information augmentation and filtering mechanism using external online sources. First, we enhance queries by appending category-specific keywords (e.g., transforming "001" to "001 adult product") via instruction-tuned 0.6B models, improving retrieval relevance from online searches that might otherwise yield irrelevant results (e.g., cars or music for "001"). This step ensures broader, contextually accurate coverage.

Subsequently, we filter retrieved information to eliminate noise: initial relevance scoring removes low-correlation content, followed by a lightweight model trained on a multi-category dataset for fine-grained consistency checks. As illustrated in Table~\ref{tab:table1}, this prevents "overthinking" in larger models, which can inflate latency and errors. Post-filtering, we extract core elements—alternative expressions, keywords, topics, hierarchies, and intent judgments—restructuring them into concise inputs for downstream LLM processing, thus enhancing efficiency and intent fidelity.

\subsubsection{Expert Rule Guidance}
Relevance criteria must adapt dynamically to inferred user intents, as static rules falter in diverse scenarios. For generic queries like "001" (spanning adult products across brands), brand matching is deprioritized; conversely, for specifics like "Yunnan Baiyao Aerosol", brand and dosage form are pivotal, disqualifying mismatches like "Yunnan Baiyao Ointment". 

Our RAG expert rule recall module addresses this by identifying entities (e.g., brand, dosage form, ingredients, region) post-query rewriting. Matched entities trigger tailored rules from a medical knowledge base, enabling fine-grained, scenario-specific assessments. This integration with LLMs ensures personalized relevance, boosting adaptability in high-volume environments like Online Medical Delivery Platforms's billions of daily queries.

\subsubsection{Cross-Modal Information Matching}
Product name standardization is fraught with inconsistencies, exacerbated by merchant manipulations for traffic gains. Text alone is easily altered, so we incorporate multimodal verification using product images to detect cheating, such as misleading names. 

We extract key attributes (name, specifications, type, users, functions) from images via multimodal models, cross-matching them against textual SPUs to normalize descriptions and flag discrepancies. This enhances robustness, preventing hijacking and ensuring reliable inputs for final relevance scoring.

\subsection{Knowledge Distillation Scheme}
To enable efficient deployment, we distill knowledge from large teacher models to compact student models using accumulated real-world data for training. This scheme balances performance with low latency and cost in industrial settings. We fine-tune Qwen3-0.6B as the student model, distilling from a Qwen3-32B teacher. The teacher generates diverse inferences via multi-prompt voting for stable labels, which the student mimics through supervised fine-tuning. This yields performance comparable to 70B-scale models in a fraction of the size, reducing inference costs while preserving accuracy in relevance assessments.

We curate high-confidence samples from Online Medical Delivery Platforms's online traffic, including good cases (accurate matches) and bad cases (misclassifications or hallucinations). These are sampled from billions of daily queries, focusing on imbalanced distributions and edge scenarios. Multi-expert validation refines labels, creating a robust dataset for distillation that informs offline iterations and bridges real-world gaps. 
In pharmaceutical search and recommendation relevance tasks, online inference and offline evaluation serve distinct yet complementary roles. Online systems must respond to user requests in extremely short time frames (typically at the k×10 ms level), imposing strict constraints on inference latency. This limits model size and usable information, creating performance ceilings for handling complex pharmaceutical scenarios, long texts, or multimodal data, and hindering global optimality.

Conversely, offline evaluation acts as a "supervisor" for relevance tasks. Using the benchmark, we leverage comprehensive search logs, real user interactions, and expert annotations to build high-quality evaluation sets. Without latency or resource constraints, offline environments allow flexible deployment of large pre-trained models, multimodal integration (e.g., text, images, structured attributes), and domain-specific fine-tuning. This enables thorough assessment of online models across sub-scenarios, uncovering long-tail issues and risks elusive to online systems.

Crucially, the accumulated high-quality samples and diagnostic data from offline evaluations provide a foundation for ongoing online model optimization. These can train and distill efficient lightweight models, allowing online systems to approximate offline performance levels. Additionally, offline results offer targeted improvements for specific drugs, diseases, and indications, fostering intelligent, refined development of pharmaceutical search and recommendation systems. 
In essence, the offline benchmark not only functions as the "gold standard" for evaluation but also drives continuous advancement of online systems, striking a balance between effectiveness and efficiency.

\begin{table}[t]
    \centering
    \caption{Category Distribution in the Benchmark}
    \begin{tabular}{l c}
        \toprule
        \textbf{First Category Name} & \textbf{Samples} \\
        \midrule
        Chinese and Western Medicines & 2820 \\
        Medical Devices (Pharmacy) & 629 \\
        Beauty and Personal Care (Pharmacy) & 241 \\
        Nutrition and Health Products & 195 \\
        Adult Products (Pharmacy) & 180 \\
        Health and Wellness & 72 \\
        Others & 263 \\
        \bottomrule
    \end{tabular}
    \vspace{-10pt}
    \label{tab:Distribution of Category Counts on the Benchmark}
\end{table}

\begin{algorithm}[htbp]
\caption{Hierarchical Relevance Judgment in AR-Med}
\label{alg:relevance_judge}
\begin{algorithmic}[1]
\REQUIRE User query, SPU (text + images)
\ENSURE $\langle$result$\rangle$level$\langle$/result$\rangle$
\STATE rewrite $\gets$ Rewrite(query)  \COMMENT{Query expansion and category tagging}
\STATE entities $\gets$ NER(rewrite)  \COMMENT{Extract core name, category, efficacy, etc.}
\STATE spu\_info $\gets$ ParseSPU(spu) \COMMENT{Brand, efficacy, dosage form, etc.}
\IF{spu.name fully contains query \textbf{and} efficacy matches}
    \RETURN $\langle$result$\rangle$Highly Relevant$\langle$/result$\rangle$
\ELSIF{all chars of query in spu.name \textbf{and} efficacy matches}
    \RETURN $\langle$result$\rangle$Highly Relevant$\langle$/result$\rangle$
\ELSIF{core name match \textbf{and} efficacy match \textbf{and} category match}
    \RETURN $\langle$result$\rangle$Highly Relevant$\langle$/result$\rangle$
\ELSIF{efficacy consistent but any core attribute mismatched}
    \RETURN $\langle$result$\rangle$Moderately Relevant$\langle$/result$\rangle$
\ELSIF{efficacy similar or complementary usage exists}
    \RETURN $\langle$result$\rangle$Weakly Relevant$\langle$/result$\rangle$
\ELSE
    \RETURN $\langle$result$\rangle$Irrelevant$\langle$/result$\rangle$
\ENDIF
\end{algorithmic}
\end{algorithm}

\begin{table*}[t]
\centering
\caption{Results of different LLMs within the proposed framework on the LocalQSMed benchmark. Here, "ALL" denotes the entire benchmark, "HARD" refers to the challenging subset, and "w/o FAILED" indicates results after excluding failed examples.}
\setlength{\tabcolsep}{1pt} %
\resizebox{\textwidth}{!}{
\begin{tabular}{l c ccc ccc ccc ccc}
\specialrule{1.2pt}{0pt}{0pt}
\multirow{2}{*}{\textbf{LLM}} 
    & \multirow{2}{*}{\textbf{Loss Rate}}
    & \multicolumn{3}{c}{\textbf{Highly Relevant}} 
    & \multicolumn{3}{c}{\textbf{Less Relevant}} 
    & \multicolumn{3}{c}{\textbf{Irrelevant}} 
    & \textbf{ALL} & \textbf{w/o FAILED} & \textbf{HARD} \\
\cmidrule(lr){3-5} \cmidrule(lr){6-8} \cmidrule(lr){9-11} \cmidrule(lr){12-14}
    & 
    & Precision & Recall & F1 
    & Precision & Recall & F1 
    & Precision & Recall & F1 
    & Accuracy & Accuracy & Accuracy \\
\midrule
\rowcolor{lightgray}
Original System & 0.00\% & \textbf{0.9760} & 0.7031 & 0.8174 & 0.0756 & 0.5984 & 0.1343 & 0.0753 & 0.2245 & 0.1128 & 0.6928 & - & - \\
Qwen3-0.6B & 6.40\% & 0.9307 & 0.8227 & 0.8734 & 0.0877 & 0.0246 & 0.0385 & 0.1333 & 0.0357 & 0.0563 & 0.7629 & 0.7867 & 0.2784 \\
Qwen3-4B   & 0.00\% & 0.9541 & \textbf{0.9741} & \underline{0.9640} & 0.2602 & 0.1475 & 0.1882 & \underline{0.4480} & 0.4706 & 0.4590 & \underline{0.9206} & \underline{0.9168} & 0.5294 \\
Qwen3-8B   & 0.00\% & 0.9553 & \underline{0.9712} & 0.9632 & \underline{0.3196} & 0.1429 & 0.1975 & 0.3647 & 0.5210 & 0.4291 & 0.9191 & \textbf{0.9190} & 0.5490 \\
Qwen3-14B  & 0.02\% & 0.9647 & 0.9594 & 0.9621 & 0.2886 & \underline{0.1982} & 0.2350 & 0.3810 & 0.6723 & \textbf{0.4863} & 0.9148 & 0.9092 & \underline{0.6106} \\
Qwen3-32B  & 0.00\% & \underline{0.9723} & 0.9326 & 0.9520 & 0.2400 & \textbf{0.3041} & \underline{0.2684} & 0.3565 & \textbf{0.6891} & \underline{0.4699} & 0.8956 & 0.8993 & \textbf{0.6611} \\
Llama-3.2-1B & 63.67\% & 0.9116 & 0.8968 & 0.9041 & 0.0270 & 0.0235 & 0.0252 & 0.0000 & 0.0000 & 0.0000 & 0.8226 & 0.8534 & 0.2913 \\
Llama-3.2-3B & 0.00\% & 0.9143 & 0.9324 & 0.9232 & 0.0217 & 0.0116 & 0.0152 & 0.0417 & 0.0217 & 0.0286 & 0.8565 & 0.8862 & 0.3358 \\
Llama-3.1-8B & 63.55\% & 0.9151 & 0.9453 & 0.9300 & 0.1591 & 0.0769 & 0.1037 & 0.0000 & 0.0000 & 0.0000 & 0.8675 & 0.8873 & 0.3504 \\
Llama-3.3-70B-Instruct & 46.86\% & 0.9296 & 0.9479 & 0.9386 & 0.0625 & 0.0472 & 0.0538 & 0.0784 & 0.0556 & 0.0650 & 0.8787 & 0.9037 & 0.3737 \\
DeepSeek-R1-Distill-Qwen-7B & 17.59\% & 0.9513 & 0.2173 & 0.3537 & 0.1622 & 0.0357 & 0.0585 & 0.3000 & 0.0303 & 0.0550 & 0.2040 & 0.1488 & 0.0699 \\
DeepSeek-R1-Distill-Qwen-14B & 0.00\% & 0.9693 & 0.9658 & \textbf{0.9658} & \textbf{0.3492} & \textbf{0.3041} & \textbf{0.3251} & \textbf{0.4569} & 0.4492 & 0.4530 & \textbf{0.9232} & 0.9125 & 0.5758 \\
DeepSeek-R1-Distill-Llama-70B & 43.68\% & 0.9293 & 0.9302 & 0.9297 & 0.0833 & 0.0541 & 0.0656 & 0.0550 & 0.0750 & 0.0635 & 0.8626 & 0.9004 & 0.3598 \\
\specialrule{1pt}{0pt}{0pt}
\end{tabular}
\vspace{-10pt}
}
\label{tab:main exp}
\end{table*}

\subsection{Benchmark: LocalQSMed}
In this section, we present the LocalQSMed dataset, constructed by randomly sampling high-frequency online data alongside error-prone cases (bad cases), resulting in a total of 4,400 query-SPU pairs. This benchmark serves as a foundational tool for offline method iteration, enabling rigorous evaluation of relevance in pharmaceutical search and recommendation scenarios. Table~\ref{tab:Distribution of Category Counts on the Benchmark} presents the detailed distribution of LocalQSMed across categories.

Real-world user queries are frequently colloquial, ambiguous, and laden with homophones or slang in medical contexts, posing significant challenges for intent disambiguation. For instance, a query like "wind-cold common cold" could refer to the disease itself or imply a search for "Wind-cold Granules" medication. Determining whether the most relevant result should be "[Quick] Compound Paracetamol and Amantadine Capsules" or "[Yunnan Baiyao] Wind-cold Granules" requires precise intent recognition and risk-aware recommendation, which is critical in medical settings to avoid potential health hazards.

SPUs are sourced from merchant-uploaded standard product names. However, merchants often engage in "traffic hijacking" through misleading nomenclature. For example, a product titled "Wuxing Jianpi Gao Pian Wan Tongren Raw Material Zhongjing Yufang Tang" exploits the popularity of "Tongrentang Jianpi Wan". Conventional machine learning and deep learning approaches are vulnerable to such manipulations due to over-reliance on keyword matching, particularly with terms like "Tongrentang". To address this, each SPU is augmented with 10-15 merchant-uploaded product images, encompassing diverse angles, ingredient lists, and actual product labels, providing multimodal cues for verification.

The LocalQSMed dataset was annotated by a team of 20 professionally trained medical experts, ensuring high-quality labels through rigorous validation. Relevance rules are dynamically adjusted based on reported bad cases and drug categories to adapt to evolving scenarios. Specifically: (1) High relevance occurs when brand, dosage form, and other key attributes are completely identical (e.g., exact matches in product specifications); (2) Low relevance (or weakly relevant) applies to cases with different brands or dosage forms but some overlapping utility; (3) Irrelevant is assigned when efficacy is completely different or the target audience diverges entirely, preventing mismatches in critical medical contexts.

%% file: sec3_experiments.tex
\section{Experiments}

\begin{table*}[htbp]
\centering
\caption{Ablation study on the proposed framework. Each row represents the progressive addition of a functional module to the base model (base (Q+S)). Abbreviations are as follows: Q (Query), S (SPU, Standard Product Unit), QI (Query Information augmentation), SI (SPU Information augmentation), IS (Internet Search), ISPI (Internet Search Page Identical check), TF (Two-Filter mechanism), CN (Common Name identification), and QC (Query Change detection).}
\resizebox{\textwidth}{!}{
\begin{tabular}{l ccc ccc ccc c}
\specialrule{1.2pt}{0pt}{0pt}
\multirow{2}{*}{\textbf{Module}} 
    & \multicolumn{3}{c}{\textbf{Highly Relevant}} 
    & \multicolumn{3}{c}{\textbf{Less Relevant}} 
    & \multicolumn{3}{c}{\textbf{Irrelevant}} 
    & \textbf{ALL} \\
\cmidrule(lr){2-4} \cmidrule(lr){5-7} \cmidrule(lr){8-10} \cmidrule(lr){11-11}
    & Precision & Recall & F1 
    & Precision & Recall & F1 
    & Precision & Recall & F1 
    & Accuracy \\
\midrule
base (Q+S) & \textbf{0.9735} & 0.9050 & 0.9380 & 0.2317 & \textbf{0.6406} & 0.3403 & 0.5926 & 0.1345 & 0.2192 & 0.8717 \\
base+QI+S & 0.9683 & 0.9386 & 0.9532 & 0.2860 & \underline{0.5760} & \textbf{0.3823} & \underline{0.7308} & 0.1597 & 0.2621 & 0.9003 \\
base+Q+SI & 0.9638 & 0.9591 & 0.9615 & 0.2832 & 0.4444 & \underline{0.3459} & \textbf{0.7500} & 0.1008 & 0.1778 & 0.9114 \\
base+QI+SI & 0.9688 & 0.9320 & 0.9500 & 0.2507 & 0.3917 & 0.3058 & 0.4510 & 0.5847 & \underline{0.5092} & 0.8966 \\
base+QI+IS+SI & 0.9576 & \underline{0.9819} & \underline{0.9696} & \underline{0.3697} & 0.2811 & 0.3194 & 0.6515 & 0.3613 & 0.4649 & \underline{0.9314} \\
base+QI+IS+ISPI+SI & 0.9620 & 0.9550 & 0.9585 & 0.2901 & 0.2166 & 0.2480 & 0.3971 & \underline{0.6807} & 0.5015 & 0.9119 \\
base+QI+IS+ISPI+TF+SI & 0.9595 & \textbf{0.9857} & \textbf{0.9725} & \textbf{0.4203} & 0.2673 & 0.3268 & 0.6905 & 0.4874 & \textbf{0.5714} & \textbf{0.9376} \\
base+QI+IS+CN+ISPI+TF+SI & 0.9675 & 0.9567 & 0.9621 & 0.2629 & 0.2120 & 0.2347 & 0.3768 & 0.6555 & 0.4785 & 0.9126 \\
base+QI+QC+CN+IS+ISPI+TF+SI &  \underline{0.9723} & 0.9326 & 0.9520 & 0.2400 & 0.3041 & 0.2684 & 0.3565 & \textbf{0.6891} & 0.4699 & 0.8956\\
\specialrule{1.2pt}{0pt}{0pt}
\end{tabular}
}
\label{tab:module_comparison}
\end{table*}

\subsection{Settings}

\paragraph{\textbf{Baselines.}}
Our primary baseline is the Original System, which was the production search relevance model in use on the Online Medical Delivery Platform prior to the implementation of AR-Med. This system is built upon a sophisticated BERT-based deep learning model, representing a strong, modern baseline that captures deep semantic relationships between user queries and product information. However, despite its capabilities, this model's knowledge is static and encapsulated entirely within its parameters. This makes it less adaptable to newly emerging medical terms or evolving pharmaceutical information without extensive and time-consuming retraining. Its performance is constrained by the knowledge present in its training data, leading to a baseline accuracy of 69.28\% on our LocalQSMed benchmark. This system serves as a strong real-world point of comparison to rigorously evaluate the significant advancements in adaptability and fact-grounding brought by our LLM-driven AR-Med framework.

\vspace{-5pt}
\paragraph{\textbf{Models.}}
For our experiments, we primarily utilize a suite of Qwen and Llama models, including Qwen3-0.6B, Qwen3-4B, Qwen2.5-7B-Instruct, Qwen2.5-VL-7B-Instruct, Qwen3-14B, Qwen2.5-VL-32B-Instruct, Qwen3-32B, Qwen2-72B-Instruct, Llama-3.2-1B, Llama-3.2-3B, Llama-3.1-8B, Llama-3.3-70B-Instruct, DeepSeek-R1-Distill-Qwen-7B, DeepSeek-R1-Distill-Qwen-14B, and DeepSeek-R1-Distill-Llama-70B. All models are deployed locally to ensure efficient and consistent evaluation. For batch inference, we employ the vLLM framework~\cite{kwon2023efficient} with 24xA100-80G GPUs, which enables high-throughput and scalable local inference.

\vspace{-5pt}
\paragraph{\textbf{Evaluation Metrics.}}
For evaluation, we focus on \textbf{accuracy}, \textbf{recall}, and \textbf{F1-score} across three relevance categories: \textbf{high relevant}, \textbf{less relevant}, and \textbf{irrelevant}. These metrics are computed based on the model's predictions on our benchmark dataset. All experiments are conducted in a controlled local environment to ensure reproducibility and fair comparison among models.

\subsection{Main Results on LocalQSMed}
We selected open-source large language models from the Qwen, Llama, and DeepSeek series to systematically evaluate our method's effectiveness on the benchmark dataset, which are shown in Table~\ref{tab:main exp}. 

The original online system achieves an accuracy of only 0.6928. Our proposed method with LLM enhancement outperforms it in most cases, with only DeepSeek-Qwen-7B failing to surpass the baseline, demonstrating the effectiveness of the approach. The best-performing models achieve accuracies approaching 0.92, representing an improvement of over 30\% compared to the original system.

Among them, the Qwen series exhibits balanced performance and stable outputs across all categories, with discrepancies from human annotations in the "Highly Relevant" category at just 3\%-5\%. Notably, while other models show lower overall Recall in "Less Relevant" categories, Qwen3-32B achieves a 68.91\% Recall in "Less Relevant." The Llama series performs slightly worse overall than Qwen. It attains high Precision and Recall in "Highly Relevant," but F1 scores in "Less Relevant" and "Irrelevant" categories are markedly lower than Qwen's. Some distilled DeepSeek models (e.g., DeepSeek-R1-Distill-Qwen-7B) experience significant sample losses, resulting in reduced accuracy and F1 scores across categories.

\subsection{Performance on Hard and Common Subsets}
Due to result losses caused by batch inference and model-specific performance issues, we introduced a new evaluation metric, w/o FAILED, which indicates results after excluding failed examples. Among the models, Qwen3-32B shows only a 2.72\% difference from human annotations in "Highly Relevant" precision, and achieves the best recall of 74.07\% in the "Less Relevant" category, making it the top-performing model with a 51.62\% improvement over the original system. Results for lossless data are provided in the appendix. 

Given the inherent sample imbalance in the benchmark, we constructed a more challenging hard dataset by applying specific rules (selecting samples with longer SPUs and shorter queries, which are prone to vendor cheating and ambiguous intent) to filter out difficult-to-classify cases, thereby achieving a balanced 1:1:1 ratio among "Highly Relevant," "Less Relevant," and "Irrelevant" samples. In this rigorous scenario, both the Qwen and Llama series demonstrated significant performance improvements as model size increased, with Qwen3-32B achieving the best overall accuracy, further validating the effectiveness of our method across different capacities. Interestingly, DeepSeek 70B performed worse than the 14B model, possibly due to differences in the underlying architecture, highlighting that, in addition to model size, the foundation and design of the model are also critical to successful online deployment. Results on the HARD dataset are provided in the appendix.

\subsection{Results Analysis Across Categories}
\begin{figure}[htbp]
    \centering
    \includegraphics[width=0.45\textwidth]{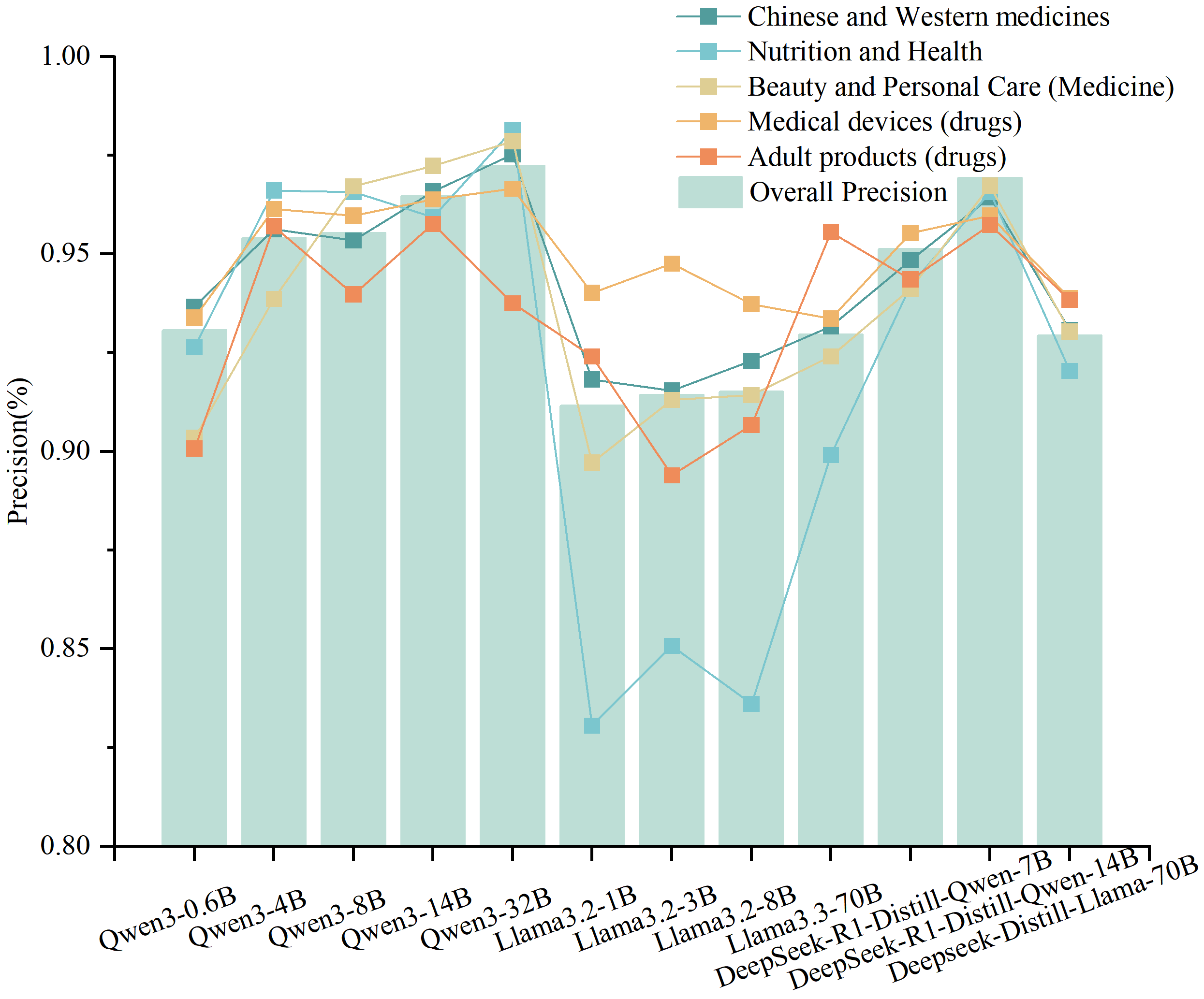}
    \caption{Per-Category results on highly relevant subset.}
    \vspace{-5pt}
    \label{fig:Categories on Highly Relevant.}
\end{figure}

To investigate the impact of different product categories on relevance prediction, we evaluated the model’s performance based on primary product classifications, as illustrated in Figure~\ref{fig:Categories on Highly Relevant.}. The bar chart shows the overall precision of the model, while the line graph depicts accuracy variations across various product categories. In total, nearly 20 categories are covered, encompassing a diverse range of pharmaceutical and related products, such as over-the-counter drugs, prescription medications, medical devices, nutritional supplements, and adult health products. To highlight key trends and avoid overly complex visualizations, we focused on the top five categories by sample volume. Detailed numerical results for all categories, including breakdowns of Precision, Recall, and F1 scores, are provided in the appendix for comprehensive reference.

Overall, the accuracy trends across different categories are consistent with the global accuracy, with only minor deviations, indicating that our method is effective and stable across different product types. For example, Qwen3-32B achieves recall rates exceeding 90\% in high-volume categories such as “Chinese and Western Medicines,” “Medical Devices,” and “Adult Products,” demonstrating strong robustness in handling ambiguous queries common in these domains, such as colloquial expressions or homophones. Even in challenging categories like “Adult Products,” where user intent can be highly implicit (e.g., numeric product codes like “001”), the method still performs stably, reflecting strong generalization capabilities. This consistency suggests that the RAG framework and knowledge distillation effectively mitigate category-specific biases, such as differences in terminology complexity or merchant manipulation, ensuring reliable relevance judgments in real-world pharmaceutical search scenarios.

\subsection{Ablation Study}
In this section, we conducted a series of ablation studies to systematically evaluate the specific contributions of each functional module to the overall performance of our model. Results are presented in Table~\ref{tab:module_comparison}. Starting from the baseline model (base, i.e., Q+S module), we progressively introduced different functional modules, including Query Info, Spu Info, Internet Search, Internet Search Page Identical, Two Filter, Common Name, and Query Change. We then performed a comprehensive comparative analysis on Precision, Recall, and F1 metrics across three categories (Highly Relevant, Less Relevant, Irrelevant) as well as the overall performance (ALL-Accuracy).

As shown in Table~\ref{tab:module_comparison}, the baseline model (base, Q+S) achieved good results in the Highly Relevant category, with Precision and Recall reaching 0.9735 and 0.9050, respectively. However, its performance on Less Relevant and Irrelevant categories was relatively poor, with F1 scores of only 0.3403 and 0.2192, indicating limitations in distinguishing more challenging samples.

As additional modules such as QI and SI were introduced, the model's performance steadily improved. For example, after adding the QI module, the F1 score for the Highly Relevant category increased from 0.9380 to 0.9532, and the F1 score for the Irrelevant category also improved, demonstrating the module's effectiveness in enhancing discriminative ability. With the further inclusion of SI, ISPI, and TF modules, the F1 scores for Less Relevant and Irrelevant categories continued to rise. Notably, after incorporating the TF module, the F1 score for the Irrelevant category reached 0.5714, and accuracy increased to 0.9376, representing the most significant overall improvement.

The results indicate that different combinations of modules have complementary effects on improving performance across categories. For example, the ISPI and TF modules significantly enhance Recall and F1 in the Irrelevant category, while the addition of CN and QC modules has a smaller impact on the Highly Relevant category but further strengthens the overall robustness of the model.

In summary, the ablation studies clearly demonstrate the distinct contributions of each functional module to model performance. In particular, the introduction of specific modules such as ISPI and TF greatly improves the model's discriminative ability in the Less Relevant and Irrelevant categories, strongly validating the effectiveness and necessity of our module design.

\begin{figure}[t]
    \centering
    \includegraphics[width=0.48\textwidth]{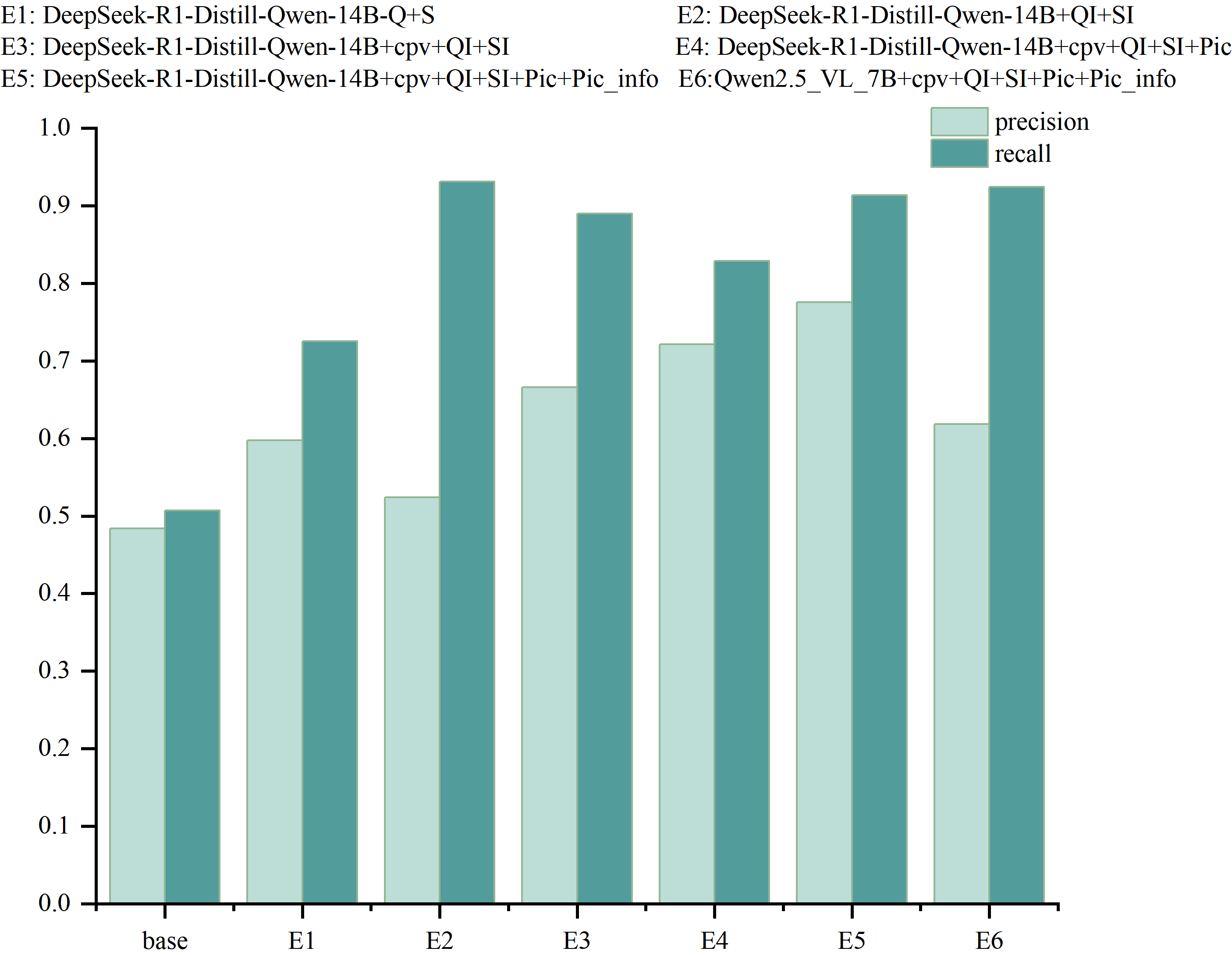}
    \caption{Results for cross-modal information matching.}
    \label{fig:Cross-Modal Information Matching.}
\end{figure}

\vspace{-2pt}
\begin{table*}[htbp]
\centering
\caption{Case Analysis 1: Cross-Modal Information Matching-SPU Consistency}
\renewcommand{\arraystretch}{1.3}
\begin{tabular}{>{\centering\arraybackslash}p{2cm}
                >{\centering\arraybackslash}p{4.7cm}
                >{\centering\arraybackslash}p{4.5cm}
                >{\centering\arraybackslash}p{5cm}}
\toprule
\textbf{Aspect} & \textbf{Standard Name} & \textbf{Image Info} & \textbf{Consistency} \\
\midrule
Brand & ``Junhong'' and ``Tongrentang'' & Only ``Junhong'' is displayed & Inconsistent: no ``Tongrentang'' \\
Product Name & ``Huoxiang Qingwei Wan Capsule'' & ``Huoxiang Qingwei Capsule'' & Mostly consistent \\
Efficacy & ``bloating,'' ``bitter taste,'' ``bad breath'' & ``Bad breath,'' ``indigestion'' & Partially consistent \\
Dosage & ``1 box'' & Not shown, but matches packaging & Consistent \\
User Group & Digestive symptoms implied & Digestive issues indicated & Consistent \\
\bottomrule
\end{tabular}
\label{tab:badcase-inconsistency}
\end{table*}

\subsection{Cross-Modal Information Matching}
On the product side, we designed multiple sets of comparative experiments to explore the optimal application position of multimodal models for product information expansion in the medical domain. As shown in Figure~\ref{fig:Cross-Modal Information Matching.}, the baseline model (base) without any expansion information achieved relatively low precision and recall, at 0.4841 and 0.5072, respectively. After introducing textual expansion (Q+S), precision increased to 0.5973 and recall to 0.7252, indicating that textual expansion significantly improves recall. With the addition of multi-dimensional expansion information such as cpv, QI, and SI, both precision and recall further improved to 0.6658 and 0.8900, demonstrating that multi-dimensional information can effectively enhance retrieval performance. When image expansion (Pic) was incorporated, precision further increased to 0.7211 and recall reached 0.8288, confirming the positive effect of multimodal expansion (image + text) on improving retrieval accuracy.

By adopting joint expansion with both spu and image, precision and recall reached their best values at 0.7759 and 0.9138, respectively, fully demonstrating that multimodal joint expansion information can significantly enhance retrieval performance. However, when all expansion information was generated solely by Qwen2.5-VL-7B, recall achieved the highest value (0.9244), but precision dropped to 0.6183, suggesting that unified expansion using a single VL model may suffer from insufficient information specificity.

The experimental results strongly demonstrate the effectiveness of multimodal expansion in improving retrieval performance on the product side. By jointly leveraging multi-source information such as text and images, not only can precision and recall be significantly enhanced, but the multi-dimensional attributes of products can also be better captured, providing solid technical support for product understanding and recommendation in real-world applications.

\begin{table*}[t]
\centering
\footnotesize
\caption{Results of Online A/B Experiments. We report the relative improvement in key metrics for our method against control groups. Abbreviations are as follows: CTR (Click-Through Rate), CVR (Conversion Rate), CXR (Click to Conversion Rate, i.e., CTR × CVR), UV (Unique Visitor), PV (Page View). AA refers to the statistical fluctuation observed in an A/A test between two identical control groups, confirming the significance of our observed improvements.}
\label{tab:global-order-summary}
\setlength{\tabcolsep}{2pt} %
\resizebox{\textwidth}{!}{
\begin{tabular}{@{} ll cccc ccc @{}}
\toprule
\multirow{2.5}{*}{Date Range} & \multirow{2.5}{*}{Scenario} & 
\multirow{2.5}{*}{Global Order Increase} & \multicolumn{3}{c}{Increase Rate / UV Metrics} & \multicolumn{3}{c}{Orders per Thousand Users / PV Metrics} \\
\cmidrule(lr){4-6} \cmidrule(lr){7-9}
& & & UV\_CXR & UV\_CTR & UV\_CVR & PV\_CXR & PV\_CTR & \\
\midrule
\multirow{4}{*}{\makecell{2025-03-30\\--\\2025-04-26}}
 & Medical Channel      & +0.29\%   & +0.08\%       & +0.21\%       & +0.01\%   & +0.58\%       & +0.51\%      \\
 &                      &           & (AA: -0.04\%) & (AA: -0.05\%) &           & (AA: -0.06\%) & (AA: +0.01\%) \\
\addlinespace[2pt]
 & Medical Main Search  & +0.35\%   & +0.16\%       & +0.19\%       & +0.43\%   & +0.59\%       & +0.33\%      \\
 &                      &           & (AA: -0.06\%) & (AA: -0.01\%) & (AA: -0.06\%) & (AA: +0.01\%) & (AA: -0.47\%) \\[-1pt]
 &                      &           &               &               &           &               & (AA: -0.53\%) \\
\addlinespace[2pt]
 & Medical Global       & +4983.0   & +0.16\%       & --            & --        & --            & --           \\
 & (AA: -1799.0 orders) & orders    & (AA: -0.06\%) &               &           &               &              \\
\bottomrule
\end{tabular}}
\end{table*}

\begin{table*}[t]
\centering
\caption{Performance Comparison of the Knowledge Distillation Scheme: Evaluating the Distilled Student Model (Qwen3-0.6B) against Teacher and Large Scale Baselines across Relevance Categories}
\resizebox{\textwidth}{!}{
\begin{tabular}{l ccc ccc ccc c}
\specialrule{1.2pt}{0pt}{0pt}
\multirow{2}{*}{\textbf{Module}} 
    & \multicolumn{3}{c}{\textbf{Highly Relevant}} 
    & \multicolumn{3}{c}{\textbf{Less Relevant}} 
    & \multicolumn{3}{c}{\textbf{Irrelevant}} 
    & \textbf{ALL} \\
\cmidrule(lr){2-4} \cmidrule(lr){5-7} \cmidrule(lr){8-10} \cmidrule(lr){11-11}
    & Precision & Recall & F1 
    & Precision & Recall & F1 
    & Precision & Recall & F1 
    & Accuracy\\
\midrule
Qwen3-32B (Qwen3-0.6b-Distill) & 0.9620 & 0.9550 & 0.9585 & 0.2901 & 0.2166 & 0.2480 & 0.3971 & 0.6807 & 0.5015 & 0.9119 \\
Qwen3-32B (Llama3.3-70B)  & 0.9569 & 0.7518 & 0.8420 & 0.2985 & 0.2151 & 0.2500 & 0.1878 & 0.3774 & 0.2508 & 0.7149 \\
\specialrule{1.2pt}{0pt}{0pt}
\end{tabular}
}
\label{tab:Knowledge Distillation}
\end{table*}

\subsection{Results on Knowledge Distillation}
As shown in Table \ref{tab:Knowledge Distillation}, we compared the consistency results between the fine-tuned consistency model and the 70B model, where the relevance discrimination model is uniformly Qwen3-32b. It is evident that the fine-tuned small model significantly outperforms the 70B model. The fundamental reason for this lies in the uniqueness of the pharmaceutical domain. By leveraging a portion of online data, we fine-tuned the Qwen3-0.6B model to enhance its pharmaceutical knowledge, thereby making it an expert in medical terminology. As a result, the model can accurately judge short queries and surpass the performance of the Llama3.3-70B model.

\subsection{Case study}
This section presents three representative cases: (1) an SPU cheating case in multimodal recognition, (2) a case on understanding query-to-internet search consistency after knowledge distillation, and (3) a case highlighting the enhancement of brand and efficacy understanding through expert rules. 
The first case (Table~\ref{tab:badcase-inconsistency}) focuses on verifying the consistency between the product standard name and image information. The model effectively detects inconsistencies such as brand piggybacking and exaggerated efficacy claims, demonstrating its capability to spot SPU cheating in multimodal scenarios. 
The second case (Table~\ref{tab:Bad Case Analysis 2: Knowledge Distillation--Query Consistency} in Appendix A.1) illustrates the model’s ability, after knowledge distillation, to judge the consistency between the query and internet search results. For example, when the query is “Mingliu,” the search result refers to “Mingliu Health Company,” while the actual SPU is “Mingliu Condom.” The model correctly identifies the inconsistency and avoids misjudgment. 
The third case (Table~\ref{tab:Bad Case Analysis3：Knowledge Distillation（Rules Recall）} in Appendix A.1) demonstrates the enhancement provided by expert rules in brand and efficacy discrimination. For instance, if the brand in the query differs from that in the SPU but the efficacy and target population are similar, the model can follow the rule to judge it as less relevant, improving adaptability in complex scenarios.

\vspace{-5pt}

\subsection{Online Evaluation}
We deployed the query method on Online Medical Delivery ’s A/B testing platform, randomly assigning 25\% of the traffic to the test group and setting up two control groups with the previous models. To ensure fairness, the experiment lasted 14 days to avoid fluctuations and holiday effects. We evaluated performance using CTR, CVR, and CXR. 
As shown in Table~\ref{tab:global-order-summary}, under medium- and high-frequency broad-intent queries, our approach improved UV\_CTR, UV\_CVR, and UV\_CXR. AA refers to the two control groups, used to rule out confounding factors; only when the AB improvement exceeds the AA fluctuation can we confirm the test group’s effectiveness. Higher CTR indicates more relevant and attractive products. Increased CVR reflects higher purchase intent after clicks. Improved CXR demonstrates an overall better conversion process.

%% file: sec4_relatedwork.tex
\section{Related Work}
\subsection{Medical Search and Recommendation}
In the early stages of medical search and recommendation systems, relevance pipelines relied heavily on traditional machine learning and information retrieval techniques. Methods such as TF-IDF-based retrieval models \cite{AIZAWA200345, salton1983introduction} calculated relevance by measuring term frequency-inverse document frequency, but they struggled to capture semantic relationships and contextual nuances. Collaborative filtering (CF) \cite{ricci2011recommender}, a staple in recommendation systems, leveraged user behavior data (e.g., browsing or ratings) to build user-item matrices for preference prediction. However, CF methods often underperformed in sparse data scenarios or cold-start problems, particularly in the medical domain where user queries are highly specialized and diverse \cite{10.1145/3372454.3372470}.
Knowledge graphs have proven effective in integrating heterogeneous data from different domains \cite{9277754}. For example, \cite{Kwon_2024} proposed a knowledge graph-based recommendation system that effectively integrates diverse medical data sources such as drugs, diseases, and symptoms to improve recommendation performance. Despite their effectiveness with structured data, knowledge graphs are costly to build and maintain, and they struggle to adapt dynamically to rapidly evolving medical data \cite{guo2020surveyknowledgegraphbasedrecommender}.

\subsection{Applications of LLMs in Medicine}
Recent advancements in large language models (LLMs) have transformed relevance pipelines in search and recommendation systems due to their robust semantic understanding and generation capabilities \cite{zhao2025surveylargelanguagemodels, zhang2025survey}. In the medical domain, LLMs have significantly enhanced relevance performance through natural language processing (NLP) techniques. For instance, models like GPT-4 demonstrate expert-level performance on medical question-answering tasks, achieving accuracies comparable to human specialists on benchmarks like USMLE \cite{nori2023capabilitiesgpt4medicalchallenge}. Specialized models such as HuaTuo and BiomedGPT, fine-tuned on Chinese medical knowledge, enable accurate diagnosis and knowledge retrieval \cite{wang2023huatuotuningllamamodel, Zhang_2024}. Chain-of-Thought (CoT) prompting has emerged as a key LLM technique for complex reasoning tasks, breaking down problems into manageable steps to improve answer accuracy \cite{10.5555/3600270.3602070}. In medical search, CoT can enhance query parsing and result ranking, but LLMs still face challenges in incomplete information scenarios, particularly with highly specialized medical terminology \cite{10.5555/3600270.3601883}. Also, LLMs have not yet matched the performance of clinical experts in medical question answering, especially in scenarios requiring high accuracy \cite{liévin2023largelanguagemodelsreason}.

Despite progress in traditional methods and LLMs, a holistic relevance optimization pipeline for medical search and recommendation remains absent. Traditional methods are constrained by limited semantic understanding, while LLMs, despite their strengths, are often applied to isolated tasks without integrating the full pipeline of search, recommendation, and user interaction \cite{10.1145/3511808.3557670}. The heterogeneity of medical data and the domain’s complexity further complicate the development of a cohesive pipeline \cite{10.1007/978-3-031-56066-8_1}.
This study aims to address this gap by proposing a comprehensive relevance pipeline that encompasses query parsing, semantic retrieval, personalized recommendation, and result optimization. 
While recent studies like AutoMIR \cite{li2025automireffectivezeroshotmedical} pioneer zero-shot medical retrieval and R2MED\cite{li2025r2medbenchmarkreasoningdrivenmedical} establishes a benchmark for reasoning-driven evaluation, our AR-Med framework takes a distinct, application-oriented approach. In contrast to unsupervised methods, we ground our system in LocalQSMed, an expert-annotated benchmark, to ensure the reliability required for industrial-scale medical platforms. AR-Med is engineered as an end-to-end solution that addresses practical challenges through knowledge distillation for efficiency and cross-modal validation against seller-side spam, bridging the gap between theoretical research and proven, scalable deployment.
We seek to structured methods × LLM semantics, delivering efficient and accurate medical search \& recommendation systems.

%% file: sec5_appendix.tex
\appendix
\section{Appendix}
\subsection{Experimental Table}
Table~\ref{tab:Bad Case Analysis 2: Knowledge Distillation--Query Consistency} presents a case study on query consistency judgment after knowledge distillation. The table contains fields such as Aspect, Query, Internet Search Result, Original Output, Output, and Actual SPU Related, highlighting the model's performance in practical product relevance determination.In the lower section, the Original Output is "Consistent (NO)", indicating that the original model failed to recognize the inconsistency.The Output is "Inconsistent (YES)", showing that the enhanced model successfully detected the inconsistency.The Actual SPU Related is "Mingliu Condom", indicating a significant difference between the actual product and the search result.
This case demonstrates that, after knowledge distillation and consistency enhancement, the model is better able to distinguish inconsistencies among the query, search result, and actual product, effectively avoiding misjudgments caused by information confusion.
\begin{table}[htbp]
\caption{Case Analysis 2: Knowledge Distillation-Query Consistency}
\renewcommand{\arraystretch}{1.3}
\begin{tabular}{
    >{\centering\arraybackslash}p{2.5cm} 
    >{\centering\arraybackslash}p{2cm} 
    >{\centering\arraybackslash}p{3.25cm}
}
\toprule
\textbf{Aspect} & \textbf{Query} & \textbf{Internet Search Result} \\
\midrule
Brand & Mingliu & Mingliu Health Company \\
\midrule
\textbf{Original Output} & \textbf{Output} & \textbf{Actual SPU Related} \\
\midrule
Consistent (NO) & Inconsistent(YES) & Mingliu Condom \\
\bottomrule
\end{tabular}
\label{tab:Bad Case Analysis 2: Knowledge Distillation--Query Consistency}
\end{table}

Table~\ref{tab:Bad Case Analysis3：Knowledge Distillation（Rules Recall）} presents a case analysis of relevance judgment enhanced by knowledge distillation and expert rules. The table includes five fields: Query, SPU, Rule Recall, Original Output, and Judgment, illustrating the decision-making process when the brand in the query differs from that in the SPU, but the efficacy and target population are similar.The Query is "Xiuzheng Xiaoshuan Tongluo Tablets".The SPU is "[Deji] Xiaoshuan Tongluo Tablets 1.8g 12 tablets 3 boards/box".The Rule Recall describes the expert rule: if the brand in the query is different from that in the SPU, but the efficacy and target population are similar, the sample should be judged as "Less Relevant".The Original Output shows that the model initially classified the sample as "Highly Relevant" (which is incorrect, hence marked NO).The Judgment field indicates the corrected output as "Less Relevant" (YES), demonstrating that, with the integration of expert rules, the model can accurately identify the true relevance in such complex cases.
\begin{table}[htbp]
\centering
\caption{Case Analysis 3: Knowledge Distillation-Rules Recall}
\renewcommand{\arraystretch}{1.3}
\begin{tabular}{
    >{\raggedright\arraybackslash}m{2.5cm} 
    >{\raggedright\arraybackslash}m{5cm}
}
\toprule
\textbf{Query} & Xiuzheng Xiaoshuan Tongluo Tablets \\
\midrule
\textbf{SPU} & [Deji] Xiaoshuan Tongluo Tablets 1.8g 12 tablets 3 boards/box \\
\midrule
\textbf{Rule Recall} & [Brand] If the brand in the query is different from the brand in the SPU, but efficacy and target population are similar, judge as Less Relevant... \\
\midrule
\textbf{Original Output} & Highly Relevant (NO)\\
\midrule
\textbf{Judgment} & Less Relevant (YES)\\
\bottomrule
\end{tabular}
\label{tab:Bad Case Analysis3：Knowledge Distillation（Rules Recall）}
\end{table}

Table~\ref{tab:Cross-Modal Information Matching} presents the performance of various large language models (LLMs) on the cross-modal information matching task. The table compares different models and their configurations in terms of Precision and Recall, evaluating their effectiveness in determining the consistency between product images and textual information.The base model serves as the baseline, with a precision of 0.4841 and a recall of 0.5072. After incorporating multimodal information, model performance improves significantly. DeepSeek-R1-Distill-Qwen-14B+cpv+QI+SI+Image+Info achieves the highest precision (0.7759), while DeepSeek-R1-Distill-Qwen-14B+QI+SI attains the highest recall (0.9311), followed by Qwen2.5-VL-7B+cpv+QI+SI+Image+Info (0.9244).

\begin{table}[htbp]
\centering
\caption{Evaluation of Cross-Modal Information Matching: Precision and Recall of Consistency Checks between Product Images and Text}
\label{tab:Cross-Modal Information Matching}
\resizebox{0.48\textwidth}{!}{
\begin{tabular}{lcc}
\specialrule{1.2pt}{0pt}{0pt}
\textbf{LLM} & \textbf{Precision} & \textbf{Recall} \\
\midrule
base & 0.4841 & 0.5072 \\
DeepSeek-R1-Distill-Qwen-14B-Q+S & 0.5973 & 0.7252 \\
DeepSeek-R1-Distill-Qwen-14B+QI+SI & 0.5241 & \textbf{0.9311} \\
DeepSeek-R1-Distill-Qwen-14B+cpv+QI+SI & 0.6658 & 0.8900 \\
DeepSeek-R1-Distill-Qwen-14B+cpv+QI+SI+Image & \underline{0.7211} & 0.8288 \\
DeepSeek-R1-Distill-Qwen-14B+cpv+QI+SI+Image+Info & \textbf{0.7759} & 0.9138 \\
Qwen2.5-VL-7B+cpv+QI+SI+Image+Info & 0.6183 & \underline{0.9244} \\
\specialrule{1.2pt}{0pt}{0pt}
\end{tabular}
}
\end{table}

Table~\ref{tab:Evaluation on w/o FAILED} presents the performance of various large language models (LLMs) on the w/o FAILED dataset, i.e., after excluding failed inference samples. The table reports Precision, Recall, and F1 scores for Highly Relevant, Less Relevant, and Irrelevant categories, as well as the overall accuracy. The highest and second-highest values in each column are highlighted in bold and underlined, respectively.

The results show that, as the model size increases, the Qwen series generally achieves better performance, with Qwen3-8B attaining the highest overall accuracy (0.9190), and Qwen3-32B excelling in Precision for Highly Relevant and Recall for Less Relevant categories. DeepSeek-R1-Distill-Qwen-14B performs best in Precision and F1 for the Less Relevant category. The Llama series achieves high Recall in the Highly Relevant category but lags behind in Less Relevant and Irrelevant categories.

Overall, after excluding failed samples, the Qwen series demonstrates stable and superior performance across all metrics, highlighting its strong capability for complex medical relevance judgment tasks.

\begin{table*}
\centering
\caption{Evaluation on w/o FAILED}
\label{tab:Evaluation on w/o FAILED}
\resizebox{\textwidth}{!}{
\begin{tabular}{l ccc ccc ccc c}
\specialrule{1.2pt}{0pt}{0pt}
\multirow{2}{*}{\textbf{LLM}} 
    & \multicolumn{3}{c}{\textbf{Highly Relevant}} 
    & \multicolumn{3}{c}{\textbf{Less Relevant}} 
    & \multicolumn{3}{c}{\textbf{Irrelevant}} 
    & \textbf{w/o FAILED} \\
\cmidrule(lr){2-4} \cmidrule(lr){5-7} \cmidrule(lr){8-10} \cmidrule(lr){11-11}
    & Precision & Recall & F1 
    & Precision & Recall & F1 
    & Precision & Recall & F1 
    & Accuracy \\
\midrule
Qwen3-0.6B & 0.9276 & 0.8536 & 0.8890 & 0.1000 & 0.0213 & 0.0351 & 0.2000 & 0.0370 & 0.0625 & 0.7867 \\
Qwen3-4B   & 0.9459 & \underline{0.9786} & 0.9620 & 0.2609 & 0.1277 & 0.1714 & 0.4545 & 0.3704 & 0.4082 & \underline{0.9168} \\
Qwen3-8B   & 0.9480 & 0.9774 & \underline{0.9625} & 0.2381 & 0.1064 & 0.1471 & \underline{0.5185} & \underline{0.5185} & \underline{0.5185} & \textbf{0.9190} \\
Qwen3-14B  & \underline{0.9629} & 0.9571 & 0.9600 & 0.2105 & 0.1702 & 0.1882 & 0.4634 & 0.7037 & \textbf{0.5588} & 0.9092 \\
Qwen3-32B  & \textbf{0.9728} & 0.9369 & 0.9545 & 0.2830 & \textbf{0.3191} & 0.3000 & 0.3846 & \textbf{0.7407} & 0.5063 & 0.8993 \\
Llama-3.2-1B & 0.9165 & 0.9274 & 0.9219 & 0.0227 & 0.0213 & 0.0220 & 0.0000 & 0.0000 & 0.0000 & 0.8534 \\
Llama-3.2-3B & 0.9192 & 0.9619 & 0.9401 & 0.0357 & 0.0213 & 0.0267 & 0.1667 & 0.0370 & 0.0606 & 0.8862 \\
Llama-3.1-8B & 0.9203 & 0.9619 & 0.9406 & 0.1429 & 0.0638 & 0.0882 & 0.0000 & 0.0000 & 0.0000 & 0.8873 \\
Llama-3.3-70B-Instruct & 0.9217 & \textbf{0.9810} & 0.9504 & 0.1176 & 0.0426 & 0.0625 & 0.0000 & 0.0000 & 0.0000 & 0.9037 \\
DeepSeek-R1-Distill-Qwen-7B & 0.9441 & 0.1607 & 0.2747 & 0.1000 & 0.0213 & 0.0351 & 0.0000 & 0.0000 & 0.0000 & 0.1488 \\
DeepSeek-R1-Distill-Qwen-14B & 0.9563 & 0.9643 & \textbf{0.9603} & \textbf{0.3043} & \underline{0.2979} & \textbf{0.3011} & 0.4762 & 0.3704 & 0.4167 & 0.9125 \\
DeepSeek-R1-Distill-Llama-70B & 0.9196 & 0.9798 & 0.9487 & 0.0000 & 0.0000 & 0.0000 & 0.0000 & 0.0000 & 0.0000 & 0.9004 \\
\specialrule{1.2pt}{0pt}{0pt}
\end{tabular}
}
\begin{tabular}{l}
\end{tabular}
\vspace{-5pt}
\end{table*}

\begin{table*}
\centering
\caption{Evaluation on HARD Data}
\label{tab:Evaluation on HARD Data}
\resizebox{\textwidth}{!}{
\begin{tabular}{l ccc ccc ccc c}
\specialrule{1.2pt}{0pt}{0pt}
\multirow{2}{*}{\textbf{LLM}} 
    & \multicolumn{3}{c}{\textbf{Highly Relevant}} 
    & \multicolumn{3}{c}{\textbf{Less Relevant}} 
    & \multicolumn{3}{c}{\textbf{Irrelevant}} 
    & \textbf{HARD} \\
\cmidrule(lr){2-4} \cmidrule(lr){5-7} \cmidrule(lr){8-10} \cmidrule(lr){11-11}
    & Precision & Recall & F1 
    & Precision & Recall & F1 
    & Precision & Recall & F1 
    & Accuracy \\
\midrule
Qwen3-0.6B & 0.3440 & 0.7611 & 0.4738 & 0.2727 & 0.0275 & 0.0500 & 0.5714 & 0.0357 & 0.0672 & 0.2784 \\
Qwen3-4B   & 0.4937 & \underline{0.9832} & 0.6573 & 0.3478 & 0.1345 & 0.1939 & \underline{0.7568} & 0.4706 & 0.5803 & 0.5294 \\
Qwen3-8B   & 0.4936 & 0.9748 & 0.6554 & \underline{0.5143} & 0.1513 & 0.2338 & 0.7126 & \underline{0.5210} & \underline{0.6019} & \underline{0.5490} \\
Qwen3-14B  & \underline{0.5616} & 0.9580 & \underline{0.7081} & 0.5581 & 0.2017 & 0.2963 & 0.7207 & 0.6723 & 0.6957 & 0.6106 \\
Qwen3-32B  & \textbf{0.6188} & 0.9412 & \textbf{0.7467} & \textbf{0.7368} & \textbf{0.3529} & \textbf{0.4773} & 0.6891 & \textbf{0.6891} & \textbf{0.6891} & \textbf{0.6611} \\
Llama-3.2-1B & 0.3000 & 0.9000 & 0.4500 & 0.3333 & 0.0244 & 0.0455 & 0.0000 & 0.0000 & 0.0000 & 0.2913 \\
Llama-3.2-3B & 0.3411 & 0.9565 & 0.5029 & 0.0000 & 0.0000 & 0.0000 & 0.3333 & 0.0217 & 0.0408 & 0.3358 \\
Llama-3.1-8B & 0.3411 & \textbf{0.9778} & 0.5057 & 0.8000 & 0.0851 & 0.1538 & 0.0000 & 0.0000 & 0.0000 & 0.3504 \\
Llama-3.3-70B-Instruct & 0.3750 & 0.9706 & 0.5410 & 0.2667 & 0.0690 & 0.1096 & 0.5714 & 0.0556 & 0.1013 & 0.3737 \\
DeepSeek-R1-Distill-Qwen-7B & 0.2927 & 0.1319 & 0.1818 & 0.4545 & 0.0521 & 0.0935 & 0.6000 & 0.0303 & 0.0577 & 0.0699 \\
DeepSeek-R1-Distill-Qwen-14B & 0.5374 & \underline{0.9664} & 0.6907 & 0.5211 & \underline{0.3109} & \underline{0.3895} & \textbf{0.7465} & 0.4492 & 0.5608 & 0.5758 \\
DeepSeek-R1-Distill-Llama-70B & 0.3702 & 0.9178 & 0.5276 & 0.2353 & 0.0656 & 0.1026 & 0.3750 & 0.0750 & 0.1250 & 0.3598 \\
\specialrule{1.2pt}{0pt}{0pt}
\end{tabular}
}
\end{table*}

Table~\ref{tab:Evaluation on HARD Data} presents the performance of various large language models (LLMs) on the HARD dataset, which consists of challenging samples with a balanced distribution of Highly Relevant, Less Relevant, and Irrelevant labels. The table reports Precision, Recall, and F1 scores for each category, as well as the overall accuracy, with the highest and second-highest values in each column highlighted in bold and underlined, respectively.

The results show that the Qwen series models consistently improve as the model size increases. Qwen3-32B achieves the best performance across multiple metrics, including Precision and F1 for Highly Relevant, all metrics for Less Relevant, Recall and F1 for Irrelevant, and overall accuracy. Qwen3-14B also performs well in several metrics. In contrast, the Llama and DeepSeek series show relatively weaker performance on the HARD dataset, particularly in the Less Relevant and Irrelevant categories, where Recall and F1 are generally low.

Overall, the Qwen series large models demonstrate superior generalization and discriminative ability in complex scenarios on the HARD dataset, highlighting their advantages in challenging medical relevance tasks.

\begin{table*}[htbp]
\centering
\caption{Evaluation of Different LLMs (Highly Relevant, Less Relevant, Irrelevant)}
\label{tab:llm-compare}
\resizebox{0.8\textwidth}{!}{
\begin{tabular}{l ccc ccc ccc}
\specialrule{1.2pt}{0pt}{0pt}
\multirow{2}{*}{\textbf{LLM}} 
    & \multicolumn{3}{c}{\textbf{Highly Relevant}} 
    & \multicolumn{3}{c}{\textbf{Less Relevant}} 
    & \multicolumn{3}{c}{\textbf{Irrelevant}} \\
\cmidrule(lr){2-4} \cmidrule(lr){5-7} \cmidrule(lr){8-10}
    & Precision & Recall & F1 
    & Precision & Recall & F1 
    & Precision & Recall & F1 \\
\midrule
GPT-4o & 0.9854 & 0.9110 & 0.9466 & 0.2149 & 0.6129 & 0.3209 & 0.7143 & 0.3244 & 0.4462 \\
Qwen3-32B & 0.9839 & 0.9429 & 0.9630 & 0.3061 & 0.6159 & 0.4109 & 0.6387 & 0.3281 & 0.4329 \\
QwQ-32B & 0.9784 & 0.9530 & 0.9656 & 0.3673 & 0.5497 & 0.4398 & 0.6555 & 0.2746 & 0.3870 \\
Qwen2-72B & 0.9715 & 0.9549 & 0.9631 & 0.3218 & 0.4677 & 0.3818 & 0.5546 & 0.3976 & 0.4638 \\
Llama3-70B & 0.9695 & 0.9664 & 0.9679 & 0.5281 & 0.4552 & 0.4888 & 0.2605 & 0.4627 & 0.3339 \\
Deepseek-r1-70B & 0.9689 & 0.9633 & 0.9661 & 0.3788 & 0.4310 & 0.4032 & 0.6471 & 0.3738 & 0.4741 \\
\specialrule{1.2pt}{0pt}{0pt}
\end{tabular}
}
\end{table*}

\subsection{Experimental Image}
\begin{figure*}[htbp]
    \centering
    \begin{subfigure}[t]{0.475\textwidth}
        \centering
        \includegraphics[width=\textwidth]{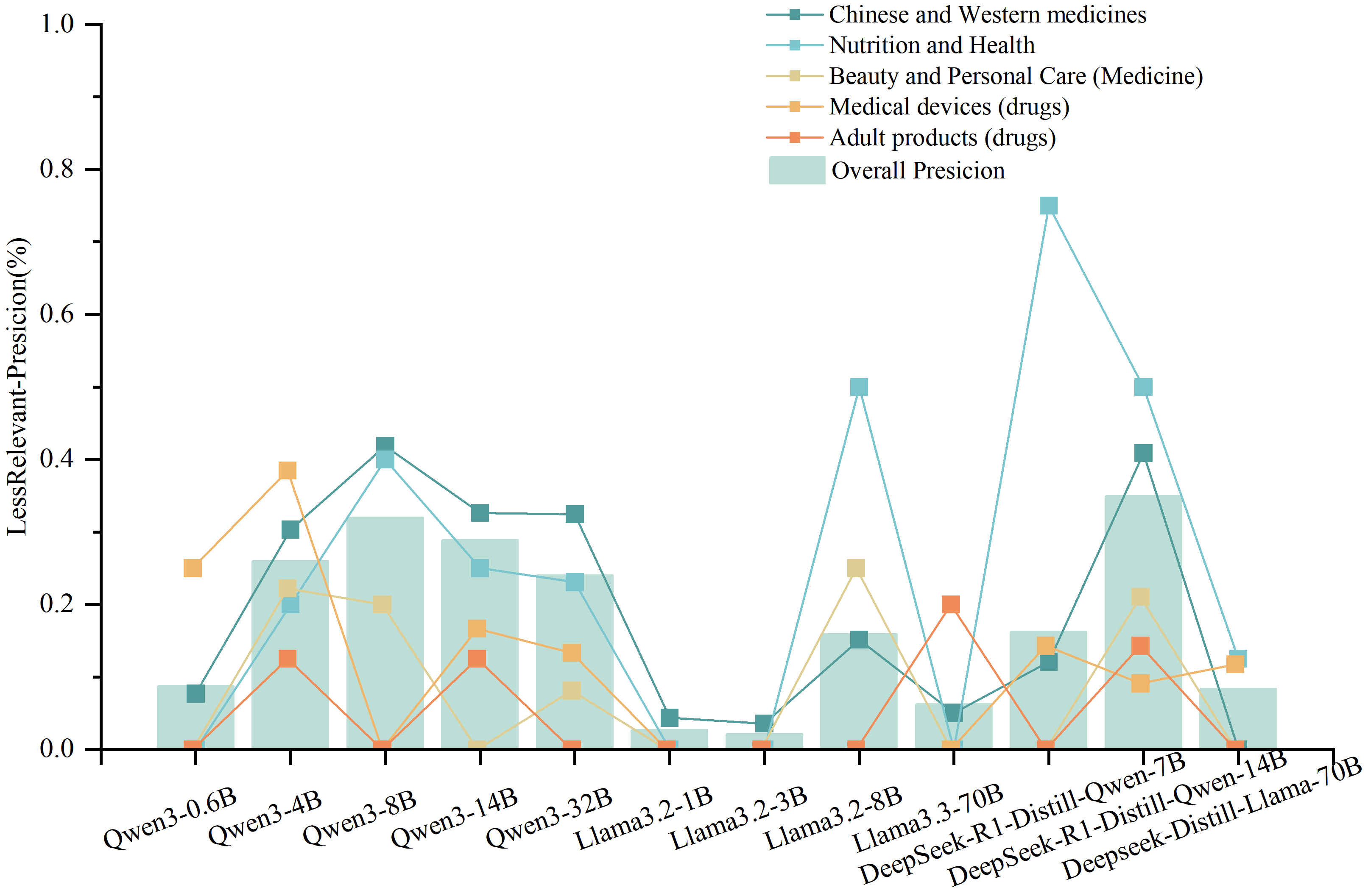}
        \caption{Precision}
        \label{fig:less-pre}
    \end{subfigure}
    \hfill
    \begin{subfigure}[t]{0.475\textwidth}
        \centering
        \includegraphics[width=\textwidth]{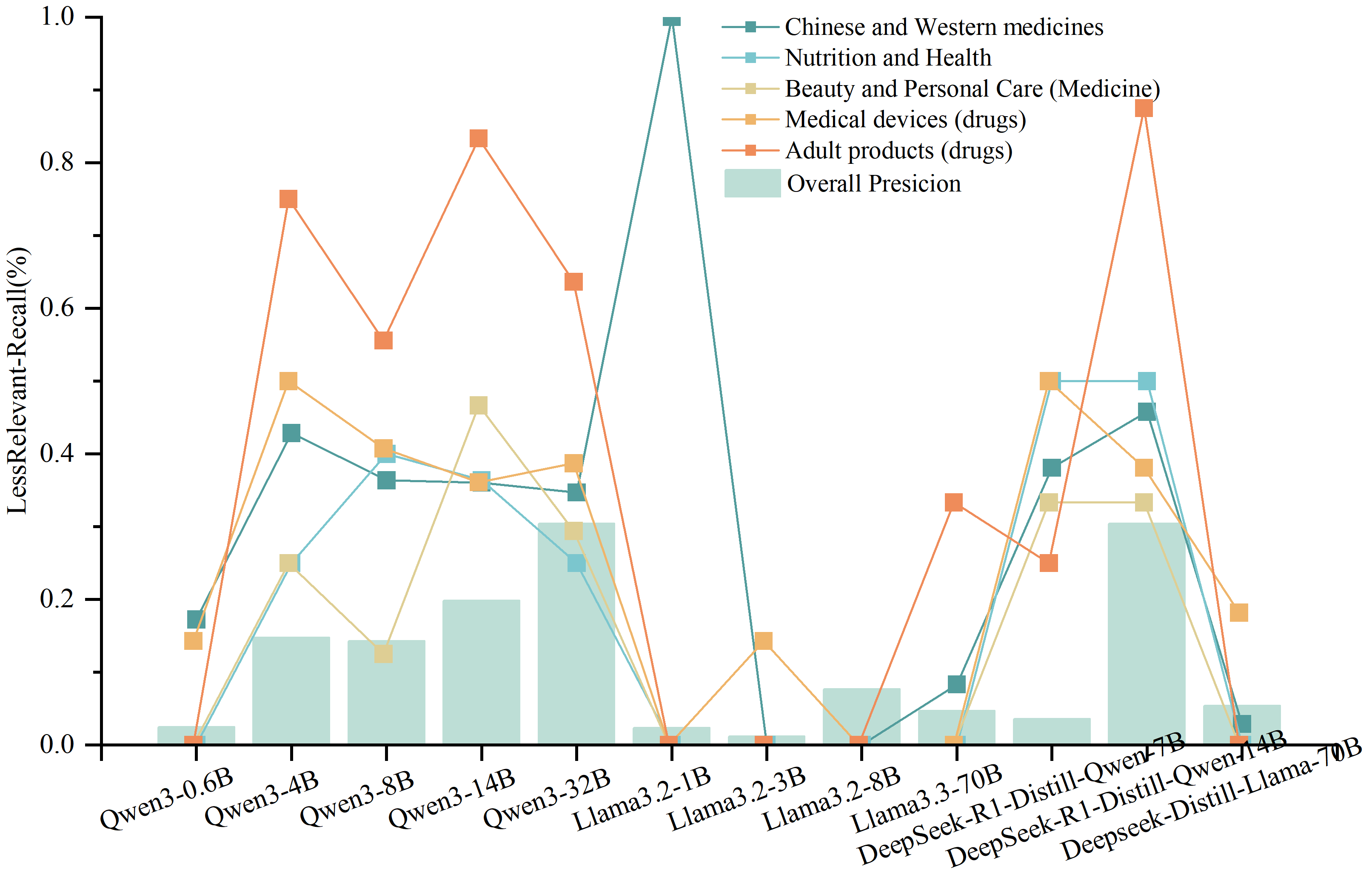}
        \caption{Recall}
        \label{fig:less-recall}
    \end{subfigure}
    \caption{Categories on Less Relevant}
    \label{fig:Less Relevant}
\end{figure*}

\begin{figure*}[htbp]
    \centering
    \begin{subfigure}[t]{0.45\textwidth}   %
        \centering
        \includegraphics[width=\textwidth]{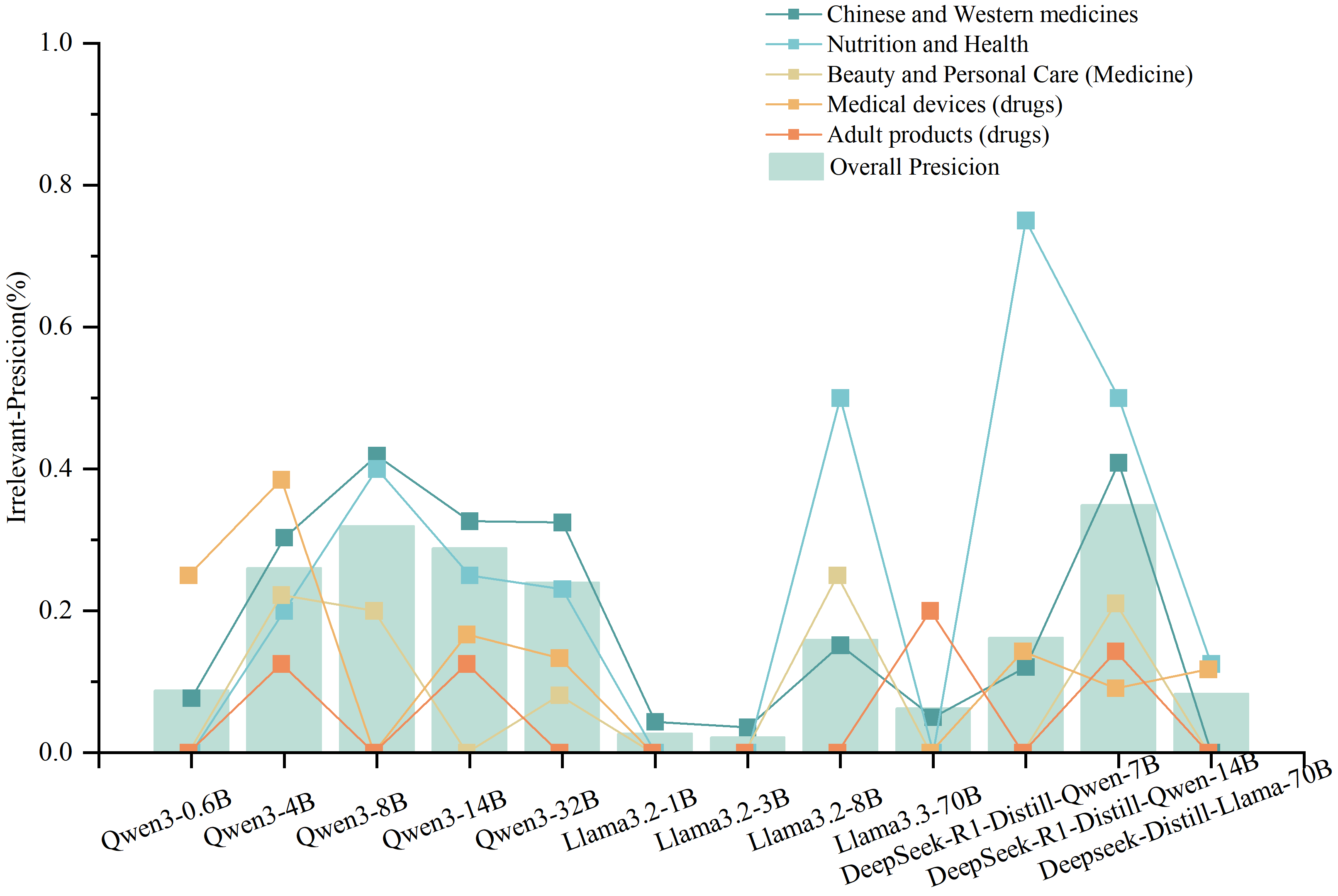}
    \end{subfigure}%
    \hspace{0.06\textwidth}%
    \begin{subfigure}[t]{0.45\textwidth}
        \centering
        \includegraphics[width=\textwidth]{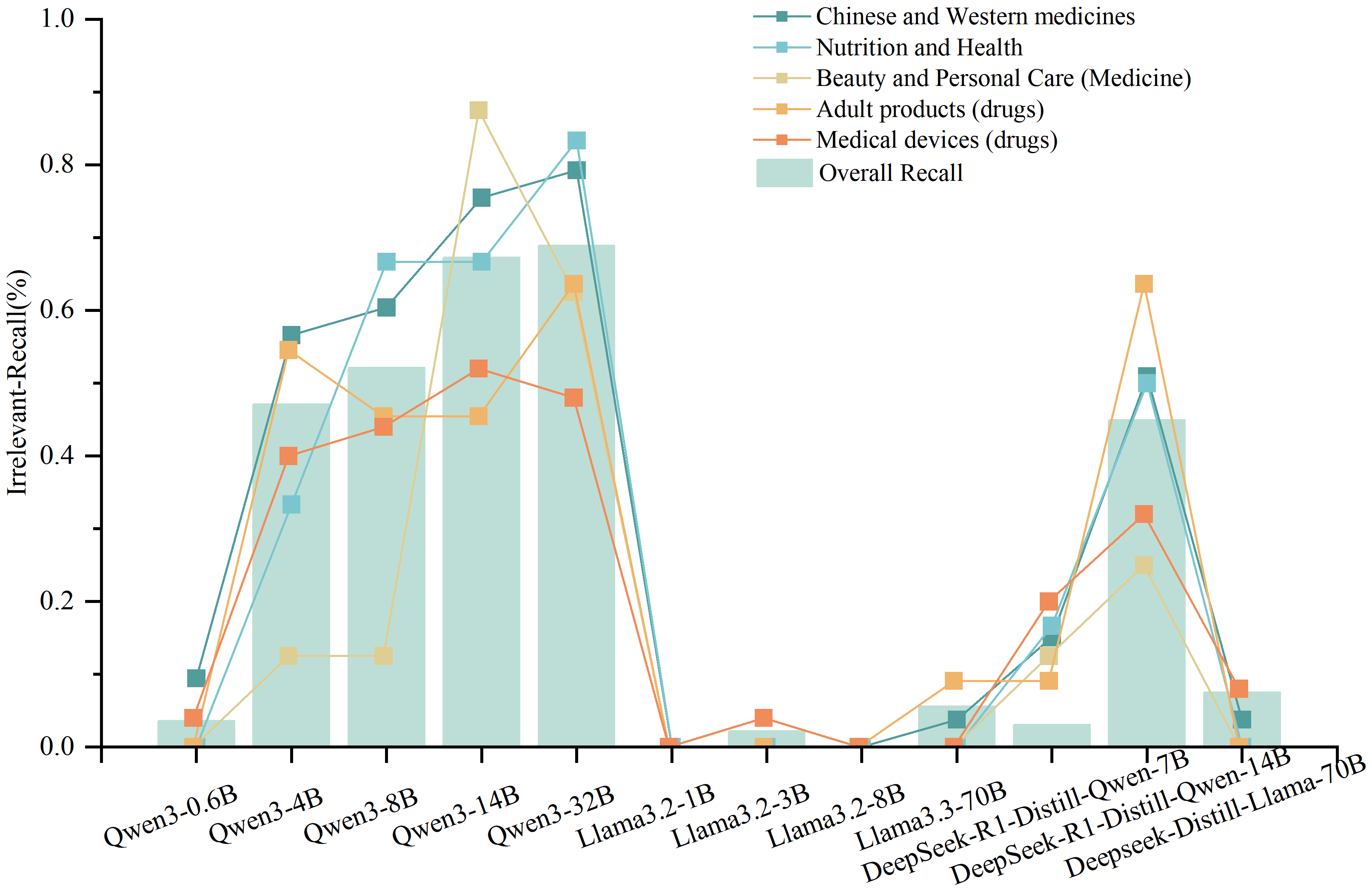}
    \end{subfigure}

    \caption{Categories on Irrelevant: (a) Precision, (b) Recall}
    \label{fig:Irrelevant Results}
\end{figure*}

As shown in Figure~\ref{fig:Less Relevant} and Figure~\ref{fig:Irrelevant Results}, the line charts illustrate the precision and recall for the “Less Relevant” and “Irrelevant” categories within the top 5 categories, while the bar charts show the overall metrics. The similar trends indicate that the model’s performance in major categories closely matches the overall results, demonstrating strong generalization and stability.

\subsection{Inference Results with Multiple Voting}

This table compares the performance of major LLMs on relevance classification tasks, with metrics for Highly Relevant, Less Relevant, and Irrelevant categories. The results indicate that most models perform well on Highly Relevant samples, while their ability to distinguish Less Relevant and Irrelevant cases varies.

82.8\% of all evaluation results have unanimous votes (6 out of 6 models), indicating high consistency among models for most samples.
When all 6 votes are consistent, the disagreement rate between LLMs and human annotations is only 1.16\%, demonstrating excellent reliability. The overall disagreement rate is 7.56%
For unanimous votes, only 2.19\% of samples are non-highly relevant, while for non-unanimous votes, this rises to 33.0\%. This suggests that model disagreement is much more likely on challenging or ambiguous cases.
In summary, the multi-model voting mechanism greatly enhances reliability, with the vast majority of samples judged consistently both among models and with human annotators. Disagreement mainly occurs in difficult cases, which may benefit from further review or advanced handling.

\subsection{Online Experiment Metrics}
These three formulas define key business metrics in search and recommendation systems. CTR measures the frequency at which users click on results, reflecting the attractiveness of the recommended content. CVR measures the proportion of clicks that lead to conversions (such as orders or purchases), indicating the effectiveness of clicks in driving actual outcomes. CXR reflects the overall efficiency from impression to conversion, being the product of CTR and CVR, or simply the ratio of conversions to impressions.

\begin{center}
\begin{align*}
&\text{CTR} = \frac{\text{Number of Clicks}}{\text{Number of Impressions}} \times 100\% 
\end{align*}
\end{center}
\vspace{-5pt}
\begin{center}
\begin{align*}
&\text{CVR} = \frac{\text{Number of Conversions}}{\text{Number of Clicks}} \times 100\%
\end{align*}
\end{center}
\vspace{-5pt}
\begin{center}
\begin{align*}
&\text{CXR} = \text{CTR} \times \text{CVR} = \frac{\text{Number of Conversions}}{\text{Number of Impressions}} \times 100\%
\end{align*}
\end{center}